%% file: main.tex
\documentclass[10pt,twocolumn,letterpaper]{article}

\PassOptionsToPackage{table,dvipsnames}{xcolor}
\usepackage[pagenumbers]{iccv} 

\input{preamble}
\usepackage{cuted}
\definecolor{iccvblue}{rgb}{0.21,0.49,0.74}
\usepackage[pagebackref,breaklinks,colorlinks,allcolors=iccvblue]{hyperref}
\usepackage[all]{hypcap}
\usepackage[accsupp]{axessibility}

\def\paperID{14005} 
\def\confName{ICCV}
\def\confYear{2025}

\title{BridgeDepth: Bridging Monocular and Stereo Reasoning with Latent Alignment}

\author{Tongfan Guan\textsuperscript{1} \quad Jiaxin Guo\textsuperscript{1} \quad  Chen Wang\textsuperscript{2} \quad  Yun-Hui Liu\textsuperscript{1}\thanks{Corresponding author}\\
$^1${The Chinese University of Hong Kong} \quad $^2${\href{https://sairlab.org}{\color{black}Spatial AI \& Robotics Lab}, University at Buffalo}\\
{\tt\small \{tfguan,jxguo,yhliu\}@mae.cuhk.edu.hk  \quad chenw@sairlab.org}
\vspace{-0.7cm}
}

\begin{document}
\maketitle

\begin{strip}
    \centering
    \includegraphics[width=0.99\linewidth]{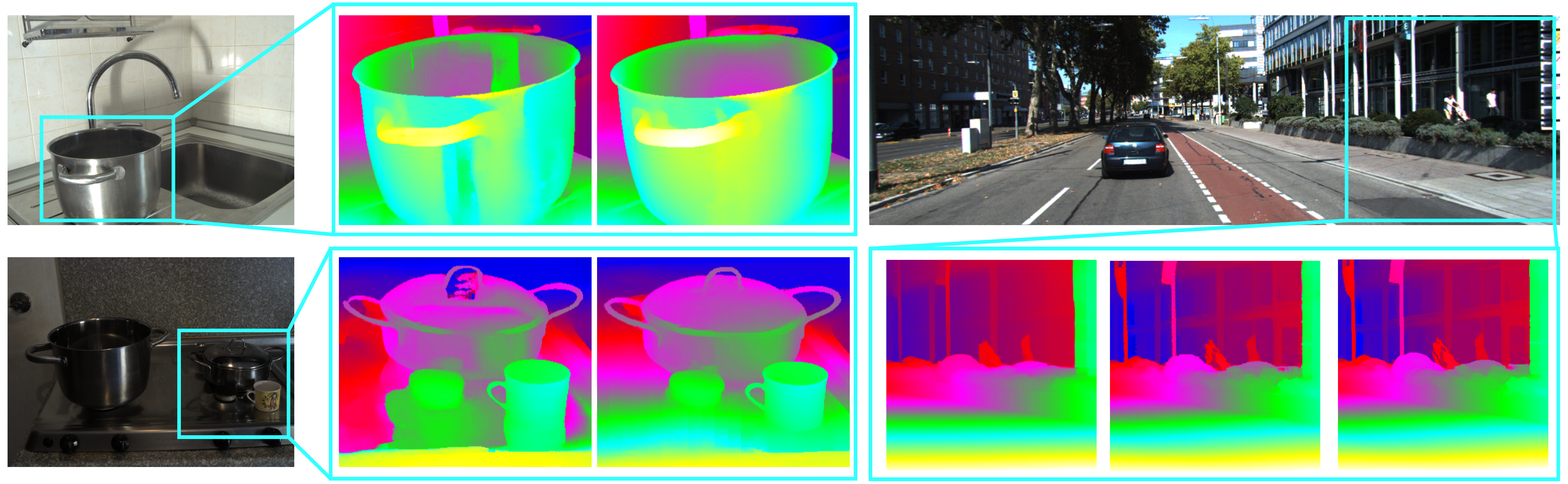}\\
    \makebox[0.21\linewidth]{}
    \makebox[0.16\linewidth]{\footnotesize \textsf{NMRF~\cite{guan2024neural}}}
    \makebox[0.17\linewidth]{\footnotesize \hspace{-0.5em}\textsf{Ours-\emph{Stereo}}}
    \makebox[0.14\linewidth]{\footnotesize \textsf{DepthAnythingV2~\cite{yang2024depth}}}
    \makebox[0.14\linewidth]{\footnotesize \textsf{Ours-\emph{Mono}}}
    \makebox[0.14\linewidth]{\footnotesize \textsf{Ours-\emph{Stereo}}}
    \vspace{-5pt}
    \captionof{figure}{Our method bridges the gap between monocular geometry reasoning and stereo pixel matching through latent representation alignment, achieving superior depth accuracy and finer details compared to the monocular model (DepthAnythingV2~\cite{yang2024depth}), while also surpassing the stereo method (NMRF~\cite{guan2024neural}) in handling textureless surfaces and reflective regions with greater robustness.}
    \label{fig:tsear}
    \vspace{-5pt}
\end{strip}

\input{sec/0_abstract}    
\input{sec/1_intro}
\input{sec/2_relatedworks}
\input{sec/3_Approach}
\input{sec/4_experiment}
\input{sec/5_conclusion}

\newpage

{
    \small
    \bibliographystyle{ieeenat_fullname}
    \bibliography{main}
}

\newpage
\input{sec/X_suppl}

\end{document}

%% file: preamble.tex
\usepackage{nicefrac}
\usepackage[super]{nth}
\usepackage{multirow}
\usepackage[table,dvipsnames]{xcolor}
\usepackage{bm}

%% file: sec/0_abstract.tex
\begin{abstract}
Monocular and stereo depth estimation offer complementary strengths: monocular methods capture rich contextual priors but lack geometric precision, while stereo approaches leverage epipolar geometry yet struggle with ambiguities such as reflective or textureless surfaces.
Despite post-hoc synergies, these paradigms remain largely disjoint in practice.
We introduce a unified framework that bridges both through iterative bidirectional alignment of their latent representations.
At its core, a novel cross-attentive alignment mechanism dynamically synchronizes monocular contextual cues with stereo hypothesis representations during stereo reasoning.
This mutual alignment resolves stereo ambiguities (e.g., specular surfaces) by injecting monocular structure priors while refining monocular depth with stereo geometry within a single network.
Extensive experiments demonstrate state-of-the-art results: \textbf{it reduces zero-shot generalization error by $\!>\!40\%$ on Middlebury and ETH3D}, while addressing longstanding failures on transparent and reflective surfaces.
By harmonizing multi-view geometry with monocular context, our approach enables robust 3D perception that transcends modality-specific limitations.
Codes available at \url{https://github.com/aeolusguan/BridgeDepth}.
\end{abstract}

%% file: sec/1_intro.tex
\section{Introduction}
\label{sec:intro}

Depth estimation is a fundamental task in 3D scene understanding, enabling diverse applications from autonomous systems to augmented reality.
Advances in deep neural networks~\cite{he2016deep,dosovitskiy2020image} have spurred two primary estimation paradigms: monocular depth estimation (MDE) and stereo matching.
Monocular methods~\cite{ranftl2020towards,ke2023repurposing,depth_anything_v1,piccinelli2024unidepth} predict depth from a single image by leveraging rich contextual priors (\eg, object semantics, scene layout).
Stereo approaches~\cite{vzbontar2016stereo,kendall2017end,lipson2021raft,guan2024neural}, in contrast, recover metric geometry through pixel correspondences across calibrated image pairs, grounded in epipolar geometry.


These paradigms rely on different cues and exhibit complementary strengths. 
MDE excels at holistic reasoning, generating plausible depth predictions even in textureless or reflective regions but suffers from inherent geometric inaccuracies due to the ill-posed nature of single-view inference.
Stereo methods achieve high metric accuracy through two-view matching but are vulnerable to correspondence ambiguities caused by occlusions, textureless regions, or specular reflections.
Critically, despite their complementarity, existing solutions treat these tasks disjointly, employing isolated pipelines that fail to unify contextual reasoning with geometric matching.
This gap fundamentally limits robustness in unconstrained scenarios where both contextual priors and geometric fidelity are indispensable.

To bridge this gap, we introduce \textbf{BridgeDepth}, a unified framework that harmonizes \textit{contextual reasoning} and \textit{geometric matching} via iterative latent representation alignment.
At the core of our method is a cross-attentive transformer module that dynamically synchronizes monocular contextual features (encoding object semantics and scene structure) with stereo hypothesis volumes (representing correspondence probabilities).
This bidirectional exchange enables monocular priors to disambiguate stereo matching under challenging conditions (e.g., specular or textureless surfaces) via learned structural constraints, while stereo geometry injects metric constraints to rectify monocular depth ambiguities.
Unlike prior work~\cite{li2024local,bae2022multi} relying on \textit{post-hoc} fusion heuristics, BridgeDepth embeds alignment directly within stereo cost aggregation, enabling the two paradigms to mutually refine their representations rather than merely complement one another. As shown in \cref{fig:tsear}, our method not only achieves consistent improvement in tackling reflective surfaces compared to stereo method NMRF~\cite{guan2024neural} but also realizes depth rectification to recover fine details compared to monocular DepthAnythingV2 (DAv2) model~\cite{yang2024depth}.

Our framework bridges monocular and stereo reasoning in two stages.
In the pre-alignment stage, the monocular branch captures contextual features of the reference stereo image, while the stereo branch prunes the disparity search space using a Disparity Proposal Network (DPN)~\cite{guan2024neural} for efficient cross-modal interaction. 
Crucially, the latent alignment stage employs our cross-attentive transformer to iteratively align monocular and stereo representations: it first queries contextual features to guide hypothesis aggregation, then updates the monocular representation using aggregated geometric constraints, enabling bidirectional refinement under dual-task supervision.
During inference, BridgeDepth produces dual outputs: relative depth~\cite{ranftl2020towards} from refined contextual features and metric disparity from stereo hypothesis representations. Comprehensive experiments demonstrate that our method achieves state-of-the-art performance across multiple benchmarks~\cite{menze2015object, geiger2012we, schops2017multi, scharstein2014high,mayer2016large}, notably reducing zero-shot error by $41.3\%$ on Middlebury~\cite{scharstein2014high} and $65.8\%$ on ETH3D~\cite{schops2017multi} versus the leading stereo baseline NMRF~\cite{guan2024neural}.
By harmonizing monocular context and stereo matching, it
transcends modality-specific limitations, enabling robust 3D perception with strong generalization.

We summarize our contributions as follows:
\begin{enumerate}
    \item BridgeDepth, a unified model jointly predicting relative depth and metric disparity via latent alignment, bridging contextual reasoning and geometric matching.
    \item A novel cross-attention transformer enabling iterative, bidirectional alignment of monocular contextual features and stereo hypothesis volume for mutual refinement.
    \item State-of-the-art performance across multiple benchmarks (KITTI, Middlebury, ETH3D, \etc), with superior efficiency, robustness, and generalization, \eg, $>\!40\%$ zero-shot error reduction on challenging datasets.
\end{enumerate}

%% file: sec/2_relatedworks.tex
\section{Related Work}
\label{sec:relatedwork}


\noindent \textbf{Monocular Depth Estimation.} Early learning-based methods~\cite{eigen2014depth,lee2019big} formulated depth estimation as direct CNN regression.
While effective in constrained settings, these approaches suffered from limited generalization due to domain gaps and inherent scale ambiguity. 
A paradigm shift emerged with MiDaS~\cite{ranftl2020towards}, which recast the task as \textit{relative scene understanding}, distilling multi-dataset priors for zero-shot inference.
Subsequent works expanded this direction: DepthAnything~\cite{depth_anything_v1} enhanced robustness through semantic feature alignment, while DAv2~\cite{yang2024depth} mitigated real-world label noise via pseudo-label distillation from a synthetic-pretrained teacher.
Alternatively, diffusion-based methods~\cite{ke2023repurposing,fu2024geowizard,he2024lotus} reimagined depth estimation as image-conditioned reconstruction---exemplified by Marigold~\cite{ke2023repurposing}, which iteratively denoises depth maps using pre-trained diffusion models to leverage generative priors for structural coherence. 
These approaches increasingly harness foundation models like DINOv2~\cite{oquab2023dinov2} and Stable Diffusion~\cite{rombach2022high} to infer plausible depth without domain-specific tuning.
Concurrently, metric depth estimation advanced through camera-aware conditioning~\cite{piccinelli2024unidepth} and canonical camera transformation~\cite{yin2023metric3d,bochkovskii2024depth}, resolving focal-distance ambiguity to enable consistent metric predictions across diverse scenes.

\smallskip\noindent \textbf{Deep Stereo Matching.} 
Traditional stereo methods relied on handcrafted similarity measures with global optimization (\eg, SGM~\cite{hirschmuller2007stereo}).
The advent of CNNs revolutionized the field: MC-CNN~\cite{vzbontar2016stereo} replaced handcrafted features with learned patch similarities, while DispNetC~\cite{mayer2016large} pioneered end-to-end disparity regression. 
Subsequent efforts formalized 3D cost volumes for disparity regression: GC-Net~\cite{kendall2017end} introduced end-to-end learning with 3D convolutions, and PSMNet~\cite{chang2018pyramid} enhanced context via spatial pyramid pooling.
Follow-up works~\cite{guo2019group,cheng2019learning,xu2022attention,shen2022pcw,Shen_2021_CVPR,Zhang2019GANet,Huang_2023_CVPR} optimized cost volume construction and aggregation.
Despite improved accuracy, high memory consumption limits their practical deployment.
RAFT-Stereo~\cite{lipson2021raft} redefined stereo matching as recurrent updates over 4D correlation volumes, enabling generalization across disparity ranges.
CREStereo~\cite{li2022practical,jing2023uncertainty} and IGEV~\cite{xu2023iterative} extended this with cascaded refinement and geometric encoding.
Recent advances like DNLR~\cite{zhao2023high}, Selective-Stereo~\cite{wang2024selective}, and Mocha-Stereo~\cite{chen2024motif} employ adaptive feature recalibration to recover high-frequency details. 
While achieving state-of-the-art accuracy, their reliance on dense cost volumes and multi-stage refinement often incurs significant computational overhead.
Another line of research~\cite{li2021revisiting,guo2022context,weinzaepfel2023croco} unleashes the power of transformer in enhancing unary feature extraction.
NMRF~\cite{guan2024neural} uniquely integrates transformers into neural Markov random fields, yielding efficient and accurate disparity estimation.

Despite progress, challenges persist in matching ambiguity and domain shifts. 
Hybrid methods~\cite{li2024local,bae2022multi} integrate monocular priors to resolve matching ambiguities but treat monocular and stereo streams separately, limiting cross-modal synergy during feature learning:
MaGNet~\cite{bae2022multi} enforces \textit{post-hoc consistency} between monocular and multi-view depth to remedy matching failures, while LoS~\cite{li2024local} propagates high-confidence disparities to ambiguous regions using monocular guidance.
In contrast, our latent alignment apparoch dynamically synchronizes monocular and stereo representations during cost aggregation to enable bidirectional refinement at the feature level.

\smallskip \noindent \textbf{Concurrent Work.}
Concurrent efforts highlight the value of integrating monocular and geometric cues.
StereoAnywhere~\cite{bartolomei2024stereo} constructs a secondary cost volume from monocular depth predictions on both stereo images and fuses it with the conventional feature-based cost volume.
DEFOM-Stereo~\cite{jiang2025defom} and MonSter~\cite{cheng2025monster} decompose stereo matching as monocular depth scale alignment and per-pixel residual refinement.
FoundationStereo~\cite{wen2025foundationstereo} adapts rich monocular priors from DAv2~\cite{yang2024depth} via a side-tuning feature network for zero-shot generalization.
Unlike these methods, our approach eliminates the need for time-consuming stereo feature extractors~\cite{bartolomei2024stereo,jiang2025defom,wen2025foundationstereo} or complex scale-alignment pipelines~\cite{jiang2025defom,cheng2025monster}.
By synchronizing monocular and stereo representations during hypothesis aggregation, rather than post-calibration or fusion, we achieve comparable or superior performance with \textbf{4$\times$} and \textbf{3$\times$} faster inference than FoundationStereo~\cite{wen2025foundationstereo} and MonSter~\cite{cheng2025monster}, respectively.

%% file: sec/3_Approach.tex
\section{Approach}

\begin{figure*}[!t]
  \centering
  \includegraphics[width=0.97\linewidth]{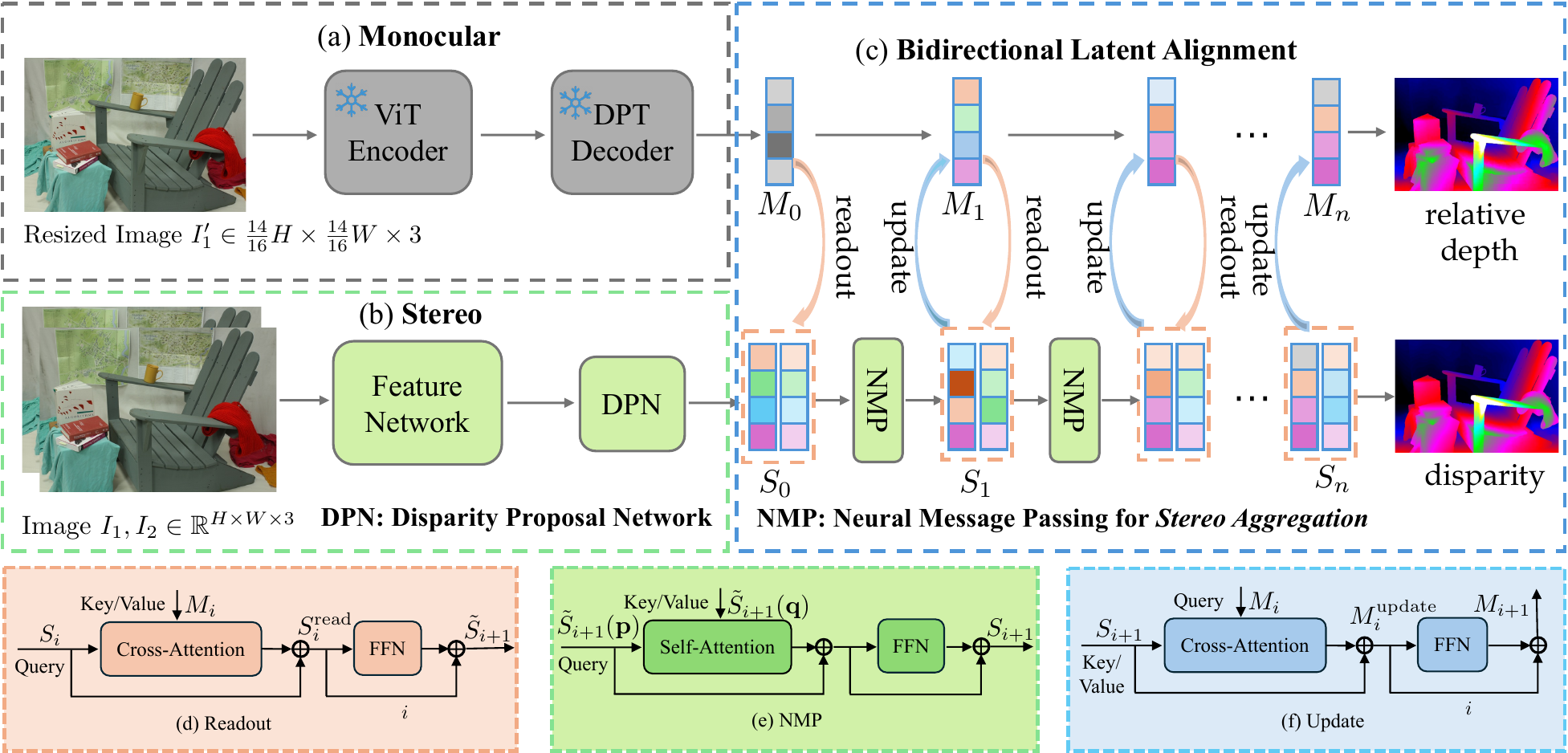}
  \caption{\textbf{Overview of the BridgeDepth model.} (1) Pre-Alignment: (a) Monocular branch extracts contextual features from the reference image; (b) stereo branch prunes disparity hypotheses (40→2) via DPN~\cite{guan2024neural}. (2) Bidirectional Latent Alignment: (c) Iteratively synchronizes monocular and stereo representations through cross-modal interaction (spatial dimensions are flattened for illustration simplicity). Outputs: relative depth (monocular) and metric disparity (stereo).}
  \label{fig:overview}
  \vspace{-0.8em}
\end{figure*}

Our approach, illustrated in~\cref{fig:overview}, operates in two stages: (1) \textbf{Pre-Alignment} for modality-specific representation preparation and (2) \textbf{Bidirectional Latent Alignment} for cross-modal synchronization.
In pre-alignment, the monocular branch extracts contextual features from the reference image, while the stereo branch employs a Disparity Proposal Network (DPN)~\cite{guan2024neural} to prune the disparity hypothesis space from 40 to 2 candidates, enabling efficient cross-modal interaction.
The core bidirectional alignment stage then iteratively synchronizes monocular contextual features with stereo hypothesis embeddings, interleaving neural message passing for hypothesis aggregation.
Finally, the aligned monocular and stereo representations are decoded into relative depth and metric disparity, respectively.

\subsection{Monocular Feature Extraction}
To inherit the generalization capability of foundation models, we leverage DepthAnythingV2 (DAv2)~\cite{yang2024depth}, a state-of-the-art monocular depth estimator pretrained on internet-scale real-world data.
As shown in~\cref{fig:overview}, DAv2 employs a DINOv2~\cite{oquab2023dinov2} encoder and a DPT~\cite{ranftl2021vision} decoder: the encoder extracts patch-wise features, while the decoder hierarchically fuses features from different depths.
To maintain compatibility with DINOv2’s 14$\times$14 patch size, we resize the reference image $\bm{I}_1$ by a factor of $14/16$, yielding $\bm{I}_1'$ with dimensions divisible by 14.
The DPT decoder outputs a fused contextual feature at the $8/14$ resolution of $\bm{I}_1'$ (equivalent to $1/2$ resolution of $\bm{I}_1$).
The feature is then downsampled via strided convolution (${\tt stride}=4$) to $1/8$ resolution, matching the spatial scale of the stereo branch.
The resulting monocular contextual feature $M_0$ has size $H/8,W/8,C$, where $H,W$ denote the original height and width of $\bm{I}_1$.

\subsection{Stereo Hypothesis Embedding}
The stereo branch encodes geometric matching cues into compact disparity hypothesis embeddings to enable efficient latent alignment with monocular contextual features.
It employs a Siamese CNN feature network to extract multi-scale features from the stereo image pair: $\bm{F}_{1,s},\bm{F}_{2,s}\in\mathbb{R}^{H/s\times W/s\times C}$, $s\in\{4,8\}$.
These hierarchical features capture correspondence patterns at progressively coarser resolutions, forming the foundation of stereo matching. 
Conventional stereo methods~\cite{kendall2017end,guo2019group} leverage these features to construct 3D cost volumes of size $D\!\times\! H/s\!\times\! W/s\!\times\! C_{\tt corr}$, where $D$ denotes the number of disparity candidates (\eg, $D=40$ for max disparity of 320 at stride $s=8$).
However, the large $D$ dimension compounds the quadratic complexity of cross-attentive monocular-stereo interaction, rendering such alignment computationally intractable.

To mitigate this, we adopt the Disparity Proposal Network (DPN)~\cite{guan2024neural}, which prunes the candidate space from $D=40$ to $K=2$ hypotheses per pixel. 
The DPN first predicts a probability distribution over all $D$ candidates using cross-correlation of features $\bm{F}_{1,8}$ and $\bm{F}_{2,8}$, then selects the top-$K$ hypotheses with the highest similarity. 
A disparity propagation module subsequently refines these hypotheses, propagating geometrically plausible candidates while discarding unlikely ones.
This significant reduction, \ie from 40 to 2, enables efficient cross-attention between monocular and stereo representations while maintaining coverage of disparity modes---the propagation module ensures retained hypotheses span the solution space to capture potential matches.
We refer details about the DPN to~\cite{guan2024neural}.

The embedding of a disparity hypothesis must integrate matching cues from both left and right views.
Given coarse-level features $\bm{F}_{1,8}, \bm{F}_{2,8}$ and the $k$-th hypothesis at pixel $(i,j)$, denoted as $\mathbf{d}_{i,j}^{k}$, corresponding embedding $S_0(i,j,k)$ is computed as:  
\begin{align}
	S_0(i,j,k)&=\text{FFN}\left(\mathbf{x}_{i,j,k}^{{\tt concat}} \parallel \mathbf{x}_{i,j,k}^{{\tt corr}}\right), \nonumber \\
	\mathbf{x}_{i,j,k}^{{\tt concat}}&=\delta_1\left(\bm{F}_{1,8}(i,j)\right) \parallel\delta_1\left(\bm{F}_{2,8}(i,j-\mathbf{d}_{i,j}^{k})\right), \label{eq:hypothesis-embedding} \\ 
	\mathbf{x}_{i,j,k}^{{\tt corr}}(g)&=\frac{G}{C}\left\langle\delta_2(\bm{F}_{1,8}(i,j,g)),\delta_2(\bm{F}_{2,8}(i,j-\mathbf{d}_{i,j}^k,g))\right\rangle \nonumber.
\end{align}
Here, $(\cdot\parallel\cdot)$ denotes channel concatenation, and $\bm{F}_{\cdot,8}(\cdot,\cdot,g)$ represent the $g^{\text{th}}$ feature group of $\bm{F}_{\cdot,8}$, with features evenly divided into $G$ groups. 
The normalization functions $\delta_1,\delta_2$ are designed to make the terms $\mathbf{x}_{i,j,k}^{{\tt concat}}$ and $\mathbf{x}_{i,j,k}^{{\tt corr}}$ share a similar data distribution.
Both $\delta_1$ and $\delta_2$ consist of a $3\times 3$ and a $1\times 1$ convolution layer, with InstanceNorm and ReLU activation following the first layer.
For sub-pixel disparities $\mathbf{d}_{i,j}^k\in\mathbb{R}$, bilinear interpolation is applied when indexing the feature map $\bm{F}_{2,8}$.

\subsection{Bidirectional Latent Alignment}
Given monocular contextual features $M_0$ and stereo hypothesis embeddings $S_0$, our bidirectional alignment module iteratively refines both representations through complementary cross-modal interactions.
The stereo hypothesis embeddings engage with monocular features in two phases: (1) Pre-aggregation readout, where cross-attention extracts global structural priors from monocular features to inform geometric matching, endowing stereo hypotheses with structure awareness, and (2) Post-aggregation update, where inverted cross-attention injects geometrically verified correspondences from the stereo volume back into the monocular stream, rectifying contextual ambiguities.
These operations, termed the \textit{monocular-readout} and \textit{monocular-update} phases,  form a self-reinforcing cycle of mutual refinement that progressively synchronizes the latent spaces over successive interaction steps---each step harmonizing semantic context with geometric constraints to achieve geometrically consistent scene understanding.

\noindent \textbf{Monocular-Readout Phase.} As illustrated in~\cref{fig:overview}(d), this phase strategically injects monocular structural priors into cost aggregation through localized cross-modal interaction. 
Specifically, prior to the $i^{\text{th}}$ aggregation step, we perform window-based cross-attention between stereo hypothesis embeddings $S_i\in\mathbb{R}^{H/8\times W/8\times K\times C}$ and monocular features $M_i\in\mathbb{R}^{H/8\times W/8\times C}$.
To manage computational complexity, both tensors are partitioned into non-overlapping $N\times N$ windows across spatial dimensions, reducing complexity from $O(HW)$ to $O(N^2)$ per embedding group.
Within each window, stereo queries $Q_s={\tt LayerNorm}(S_i){\tt W_q}$ attend to monocular keys $K_m={\tt LayerNorm}(M_i){\tt W_k}$ and values $V_m={\tt LayerNorm}(M_i){\tt W_v}$, followed by FFN feature transformation and residual fusions:
\begin{align}
	S_i^{\text{read}}&=S_i + \text{WindowAttn}(Q_s, K_m, V_m),\\
	\tilde{S}_{i+1}&=S_i^{\text{read}}+\text{FFN}(\text{LayerNorm}(S_i^{\text{read}})),
\end{align}
where FFN denotes a two-layer MLP with GELU~\cite{hendrycks2016gaussian} activation, and ${\tt WindowAttn}$ implements multi-head attention with content-adaptive positional bias~\cite{li2021revisiting,guan2024neural}.
To preserve spatial alignment, we extend standard position encoding with dual-role biases:
\begin{align}
	\text{WindowAttn}(Q, K, V)&=\text{softmax}\left(\frac{QK^T}{\sqrt{d}}+\beta\right)V,\\
    \beta&=Q^TR^k+K^TR^q.\label{eq:bias}
\end{align}
Here, $R^q,R^k\in\mathbb{R}^C$ are retrieved from a learnable position table $R\in\mathbb{R}^{2\times (2M-1)\times (2M-1)\times C}$, where the first dimension distinguishes query/key roles.
This adaptive bias modulates attention weights through both content similarity and content-position interactions.
The refined embeddings $\tilde{S}_{i+1}$, now enriched with monocular-derived semantic constraints, proceed to stereo cost aggregation.

\smallskip\noindent\textbf{Cost Aggregation.}
Building on the monocular-augmented embeddings $\tilde{S}_{i+1}$, we utilize Neural Message Passing (NMP)~\cite{guan2024neural}, a structured variant of self-attention, to perform cost aggregation.
Departing from conventional Swin-Attention~\cite{liu2021swin}, NMP decouples cost aggregation into two complementary processes. 
(1) Inter-pixel attention propagates geometric consensus across spatial neighborhoods, aggregating hypotheses from adjacent pixels within local windows to reinforce disparities that align with scene structure.
(2) Intra-pixel attention resolves hypothesis competition within individual pixels via disparity-level interactions, suppressing inconsistent candidates while amplifying geometrically plausible matches.
As shown in~\cref{fig:overview}(e), following this attentional aggregation, a feed-forward network (FFN) nonlinearly transforms the refined embeddings to enhance feature discriminability. 
The resulting $S_{i+1}$ embeddings consolidate both geometrically validated constraints and monocular structural priors, forming a cohesive representation for subsequent monocular-update phases.

\smallskip\noindent\textbf{Monocular-Update Phase.}
Completing the bidirectional alignment loop, this phase inversely propagates stereo-validated geometric evidence into monocular stream.
Mirroring the cross-attention structure of the readout phase, we treat monocular features $M_i$ as the queries and the aggregated stereo embeddings $S_{i+1}$ as keys/values.
Interactions remain constrained to local $N\times N$ windows, where $M_i$ dynamically attends to disparity-consistent patterns in $S_{i+1}$:
\begin{align}
	M_i^{\text{update}}&=M_i + \text{WindowAttn}(Q_m, K_s, V_s),\\
	M_{i+1}&=M_i^{\text{update}}+\text{FFN}(\text{LayerNorm}(M_i^{\text{update}})).
\end{align}
Here, queries $Q_m={\tt LayerNorm}(M_i){\tt W_q'}$ and keys/values $K_s$/$V_s={\tt LayerNorm}(S_{i+1}){\tt W_k'}$/${\tt W_v'}$.
The same content-adaptive positional bias mechanism (\cref{eq:bias}) preserves pixelwise alignment during reverse projection. This process injects stereo-verified constraints (e.g., occlusion boundaries, depth-ordering) into $M_i$, resolving monocular ambiguities through geometric grounding. 
The updated $M_{i+1}$ retains semantic richness while encoding latent geometric coherence, thereby closing the iterative alignment cycle.

\smallskip\noindent\textbf{Fine-Grained Detail Recovery.} 
While bidirectional alignment operates efficiently at $H/8 \times W/8$ resolution, we introduce a coarse-to-fine cascade to recover high-frequency geometric details.
After $l$ steps of alignment, the stereo embeddings are decoded into a full-resolution disparity map $D_{\text{coarse}}\in\mathbb{R}^{H\times W}$. This coarse prediction is then downsampled to $H/4\times W/4$ via edge-preserving median pooling.
The downsampled disparity initializes refined stereo embeddings $S_l\in\mathbb{R}^{H/4\times W/4\times C}$ through feature warping and grouped correlation with fine-scale backbone features $\{\bm{F}_{1,4},\bm{F}_{2,4}\}$, as formalized in~\cref{eq:hypothesis-embedding}.
\begin{table*}[!t]
\small
\centering
\begin{tabular}{l c c c c c c c c c c c c}
\toprule
& \multicolumn{4}{c}{KITTI 2012 Reflective} & \multicolumn{4}{c}{KITTI 2012} & \multicolumn{3}{c}{KITTI 2015} &\\
\cmidrule(lr){2-5}
\cmidrule(lr){6-9}
\cmidrule(lr){10-12}
Method & \multicolumn{2}{c}{Out-2} & \multicolumn{2}{c}{Out-3} & \multicolumn{2}{c}{Out-2} & \multicolumn{2}{c}{Out-3}  
& BG & FG & ALL & Runtime$^\dag$\\
\cmidrule(lr){2-3}
\cmidrule(lr){4-5}
\cmidrule(lr){6-7}
\cmidrule(lr){8-9}
& Noc & All & Noc & All & Noc &  All &  Noc &  All & \multicolumn{3}{c}{All Areas} & (s) \\
\hline
GANet-deep~\cite{Zhang2019GANet}  & 10.75 & 12.94 & 6.22 & 7.92 & 1.89 & 2.50 & 1.19 & 1.60 &1.48 &3.46 & 1.81 & - \\
CSPN~\cite{cheng2019learning} & 10.40 & 12.73 & 6.40 & 8.17 & 1.79 & 2.27 & 1.19 & 1.53 & 1.51 & 2.88 & 1.74 & -\\
RAFT-Stereo~\cite{lipson2021raft} & - & - & - & - & 1.92 & 2.42 & 1.30 & 1.66 & 1.58 & 3.05 & 1.82 & 0.38\\
LEAStereo~\cite{cheng2020hierarchical} & 9.66 & 11.40 & 5.35 & 6.50 & 1.90 & 2.39 & 1.13 & 1.45 & 1.40 & 2.91 & 1.65 & -\\
ACVNet~\cite{xu2022attention} & 11.42 & 13.53 & 7.03 & 8.67 & 1.83 & 2.35 & 1.13 & 1.47 & 1.37 & 3.07 & 1.65 & 0.2\\
IGEV-Stereo~\cite{xu2023iterative} & 7.57 & 8.80 & 4.35 & 5.00 & 1.71 & 2.17 & 1.12 & 1.44 & 1.38 & 2.67 & 1.59 & 0.18\\
PCWNet~\cite{shen2022pcw} & 8.94 & 10.71 & 4.99 & 6.20 & 1.69 & 2.18 & 1.04 & 1.37 & 1.37 & 3.16 & 1.67 & -\\
LoS~\cite{li2024local} & 6.31 & 7.84 & 3.47 & 4.45 & 1.69 & 2.12 & 1.10 & 1.38 & 1.42 & 2.81 & 1.65 & - \\
Mocha-Stereo~\cite{chen2024motif} & 6.97 & 8.10 & 3.83 & 4.50 & 1.64 & 2.07 & 1.06 & 1.36 & 1.36 & \cellcolor{yellow!25}2.43 & 1.53 & - \\
Selective-IGEV~\cite{wang2024selective} & 6.73 & 7.84 & 3.79 & 4.38 & 1.59 & 2.05 & 1.07 & 1.38 & 1.33 & 2.61 & 1.55 & 0.24 \\
NMRF~\cite{guan2024neural} & 10.02 & 12.34 & 6.35 & 8.11 & 1.59 & 2.07 & 1.01 & 1.35 & 1.28 & 3.13 & 1.59 & \cellcolor{red!25}\textbf{0.09}\\
\hline
IGEV++~\cite{xu2024igev++} & \cellcolor{red!25}\textbf{5.49} & 6.86 & \cellcolor{red!25}\textbf{2.68} & \cellcolor{yellow!25}3.48 & \cellcolor{yellow!25}1.36 & \cellcolor{yellow!25}1.74 & 0.89 & 1.13 & 1.15 & 2.80 & 1.43 & 0.48 \\
MonSter~\cite{cheng2025monster} & \cellcolor{yellow!25}5.66 & \cellcolor{yellow!25}6.81 & \cellcolor{yellow!25}2.75 & \cellcolor{red!25}\textbf{3.38} & \cellcolor{yellow!25}1.36 & 1.75 & \cellcolor{yellow!25}0.84 & \cellcolor{yellow!25}1.09 & \cellcolor{red!25}\textbf{1.13} & 2.81 & \cellcolor{yellow!25}1.41 & 0.45\\
DEFOM-Stereo~\cite{jiang2025defom} & 5.76 & \cellcolor{red!25}\textbf{6.72} & 3.04 & 3.56 & 1.43 & 1.79 & 0.94 & 1.18 & 1.25 & \cellcolor{red!25}\textbf{2.23} & \cellcolor{yellow!25}1.41 & 0.61\\
BridgeDepth (Ours) & 5.80 & 6.85 & 2.91 & \cellcolor{yellow!25}3.48 & \cellcolor{red!25}\textbf{1.32} & \cellcolor{red!25}\textbf{1.65} & \cellcolor{red!25}\textbf{0.83} & \cellcolor{red!25}\textbf{1.03} & \cellcolor{red!25}\textbf{1.13} & 2.73 & \cellcolor{red!25}\textbf{1.40} & \cellcolor{yellow!25}0.13\\
\bottomrule
\end{tabular}
\vspace{-0.5em}
\caption{\textbf{Benchmark results on KITTI 2012 Reflective, KITTI 2012, and KITTI 2015 datasets.} For KITTI 2012/Reflective, we evaluate outlier ratios (Out-$x$) for disparity errors exceeding $x$ pixels in non-occluded (Noc) and full-image (All) regions. For KITTI 2015, we quantify the D1 error rate across background (BG), foreground (FG), and combined (ALL) partitions. ($\dag$): Benchmarked on GTX 3090.}
\label{tab:results_kitti}
\vspace{-0.5em}
\end{table*}

\begin{figure*}[!t]
  \centering
  \tabcolsep=0.02cm
  \begin{tabular}{c c}
    \raisebox{1.4em}{\rotatebox{90}{\footnotesize \textsf{Ours} \hspace{2.2em} \textsf{NMRF}~\cite{guan2024neural} \hspace{1.0em} \textsf{Left Image}}} & \includegraphics[width=0.96\linewidth]{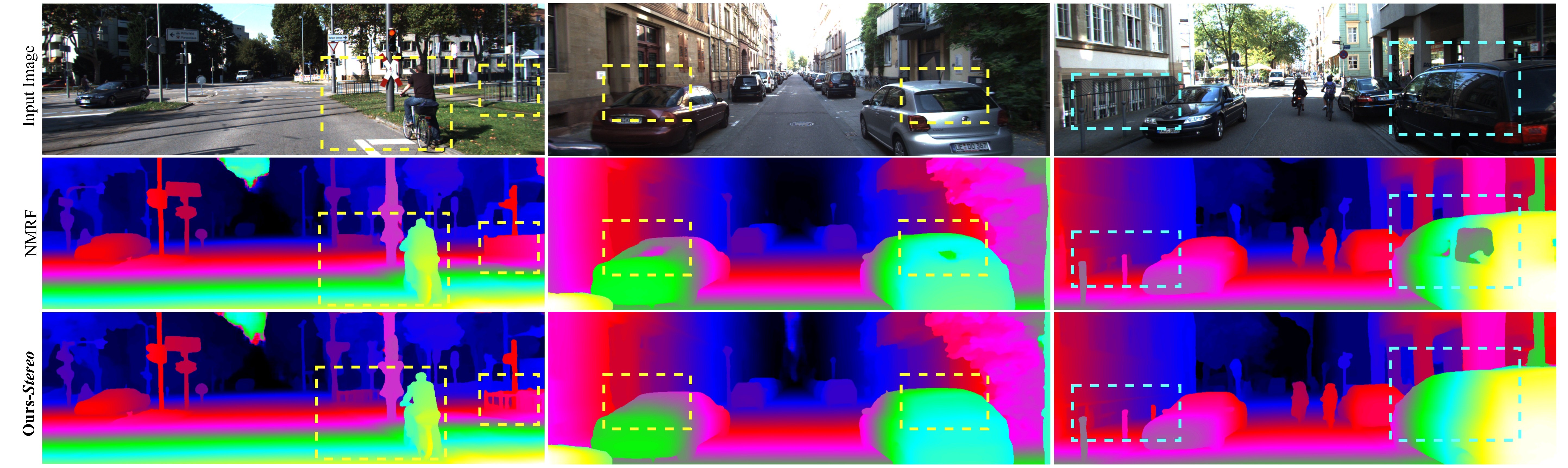} \\
  \end{tabular}
  \vspace{-0.5em}
  \caption{\textbf{Qualitative Comparison on KITTI with fine-tuned model on mixed KITTI 2012/2015 data}. We compare with the leading baseline NMRF~\cite{guan2024neural}, exhibiting high robustness and accuracy to the challenging region with transparent window and rich details.}
  \label{fig:kitti_ft}
  \vspace{-1.0em}
\end{figure*}

To preserve equivalent contextual regions between the higher-resolution stereo embeddings $S_l\in\mathbb{R}^{H/4\times W/4 \times C}$ and coarser monocular features $M_l\in\mathbb{R}^{H/8\times W/8 \times C}$, the cross-attention window size of monocular branch is halved, \ie, $N\rightarrow N/2$.
Bidirectional alignment then resumes: stereo hypotheses integrates localized structural cues from $M_l$ via reduced-window cross-attention, while the monocular stream absorbs fine-grained geometric cues from $S_l$ through the content-adaptive positional bias defined in~\cref{eq:bias}.
This asymmetric resolution coupling, \ie, processing stereo embeddings at $1/4$ resolution while retaining monocular features at $1/8$, effectively recovers thin structures and sharp occlusion boundaries without computationally intensive high-resolution processing, establishing a complete coarse-to-fine alignment framework.

\smallskip\noindent\textbf{Dual-Outputs.}
After $l'$ additional bidirectional alignment iterations, the refined representations $S_{l+l'}$ (stereo) and $M_{l+l'}$ (monocular) are decoded into dual outputs: a metric disparity map from the stereo branch and a relative depth map from the monocular branch. 
Both branches utilize a shared decoder architecture comprising a lightweight 3-layer MLP with ReLU activation, followed by a pixel unshuffle operation for full resolution reconstruction.
For the stereo branch, the MLP first projects embeddings $S_{l+l'}\in\mathbb{R}^{H/4\times W/4\times C}$ to a ${H/4\times W/4\times 16}$ feature map, which is then reassembled into a full-resolution metric disparity map $D\in\mathbb{R}^{H\times W}$ via pixel unshuffle.
The monocular branch follows an analogous process.

\subsection{Rationale of Dual-Outputs}
While the metric disparity output already provides geometrically accurate reconstruction, the co-estimated relative depth map serves two critical purposes.
First, it helps to validate the efficacy of our latent cross-modal alignment.
Consistency (see~\cref{tab:results_zero_shot} and~\cref{fig:mono_zero}) between the monocular output (initially scale-inconsistent) and ground truth depth demonstrates that monocular features are indeed being refined with geometrically verified constraints.
More fundamentally, the dual-output framework establishes a generalizable cross-modal synergy beyond stereo matching. 
Our latent alignment mechanism inherently models bidirectional relationships between geometric hypotheses and contextual priors, a paradigm transferable to tasks requiring complementary cues. 
For instance, in optical flow estimation, one could align monocular depth cues with motion hypotheses through analogous bidirectional interactions, simultaneously recovering geometrically consistent depth and temporally coherent flow. 
This positions our framework as a blueprint for tasks requiring coupled geometric-contextual reasoning, where dual outputs mutually regularize through latent alignment.
\begin{table*}[!t]
\small
\centering
\resizebox{0.98\linewidth}{!}{
\begin{tabular}{r c c c c c c c}
\toprule 
Method & RAFT-Stereo~\cite{lipson2021raft} & DLNR~\cite{zhao2023high} & 
Selective-IGEV~\cite{wang2024selective} & NMRF~\cite{guan2024neural} & Mocha-Stereo~\cite{chen2024motif} & BridgeDepth & BridgeDepth-L\\
\midrule
EPE & 0.56 & 0.48 & 0.44 & 0.45 & 0.41 & 0.37 & \textbf{0.32}\\
BP-1 &6.63 &5.39&4.98&4.50&5.07&3.67&\textbf{3.25}\\
\bottomrule
\end{tabular}
}
\vspace{-0.5em}
\caption{\textbf{Quantitative evaluation on Scene Flow test set.} Our method achieves state-of-the-art performance on both metrics. \textbf{Bold}: Best.}
\vspace{-1.2em}
\label{tab:results_sceneflow}
\end{table*}
\subsection{Loss Function}
We supervise the framework using dual losses.
For the stereo output, we use the L1 loss for $\mathcal{L}_{\text{stereo}}$, while the affine-invariant loss~\cite{ranftl2020towards} is employed as $\mathcal{L}_{\text{mono}}$ for monocular output.
Let $\{\mathbf{d}_i\}_{i=0}^{l+l'-1}$ and $\{\mathbf{m}_i\}_{i=0}^{l+l'-1}$ denote the disparity and relative depth predictions across all $l+l'$ iterations. The losses are defined as:
\begin{align}
    \mathcal{L}_{\text{stereo}} &= \sum_{i=0}^{l+l'-1} w_i ||\mathbf{d}_i - \mathbf{d}_{\text{gt}}||_1, \\
    \mathcal{L}_{\text{mono}} &= \sum_{i=0}^{l+l'-1} \gamma^{l+l'-i} \rho(\mathbf{m}_i, \mathbf{d}_{\text{gt}}),
\end{align}
where $\rho(\cdot,\cdot)$ denotes the affine-invariant loss, $\gamma=0.8$ exponentially increases the weights, ${w_i}$ balances contribution across stereo iterations, and $\mathbf{d}_{\text{gt}}$ is ground-truth disparity.

%% file: sec/4_experiment.tex
\section{Experiments}

\subsection{Implementation Details}
We implement BridgeDepth using PyTorch on NVIDIA RTX 3090 GPUs. 
The AdamW optimizer is adopted with a one-cycle learning rate scheduler across all experiments.
Data augmentations similar to~\cite{lipson2021raft} are performed.
The monocular branch employs a frozen ViT-L version of DAv2~\cite{yang2024depth}, preserving its pretrained generalization capability on real-world data.
For stereo feature extraction, we employ the $\tt BasicEncoder$ originated from RAFT~\cite{teed2020raft}.
Pretraining is conducted on the Scene Flow~\cite{mayer2016large} dataset for 300K iterations with a batch size of 8. 
Inputs are randomly cropped to $384\times 768$ resolution, and the learning rate follows a cosine one-cycle schedule peaking at 5e-4.
\begin{figure}[!t]
  \centering
  \includegraphics[width=1.0\linewidth]{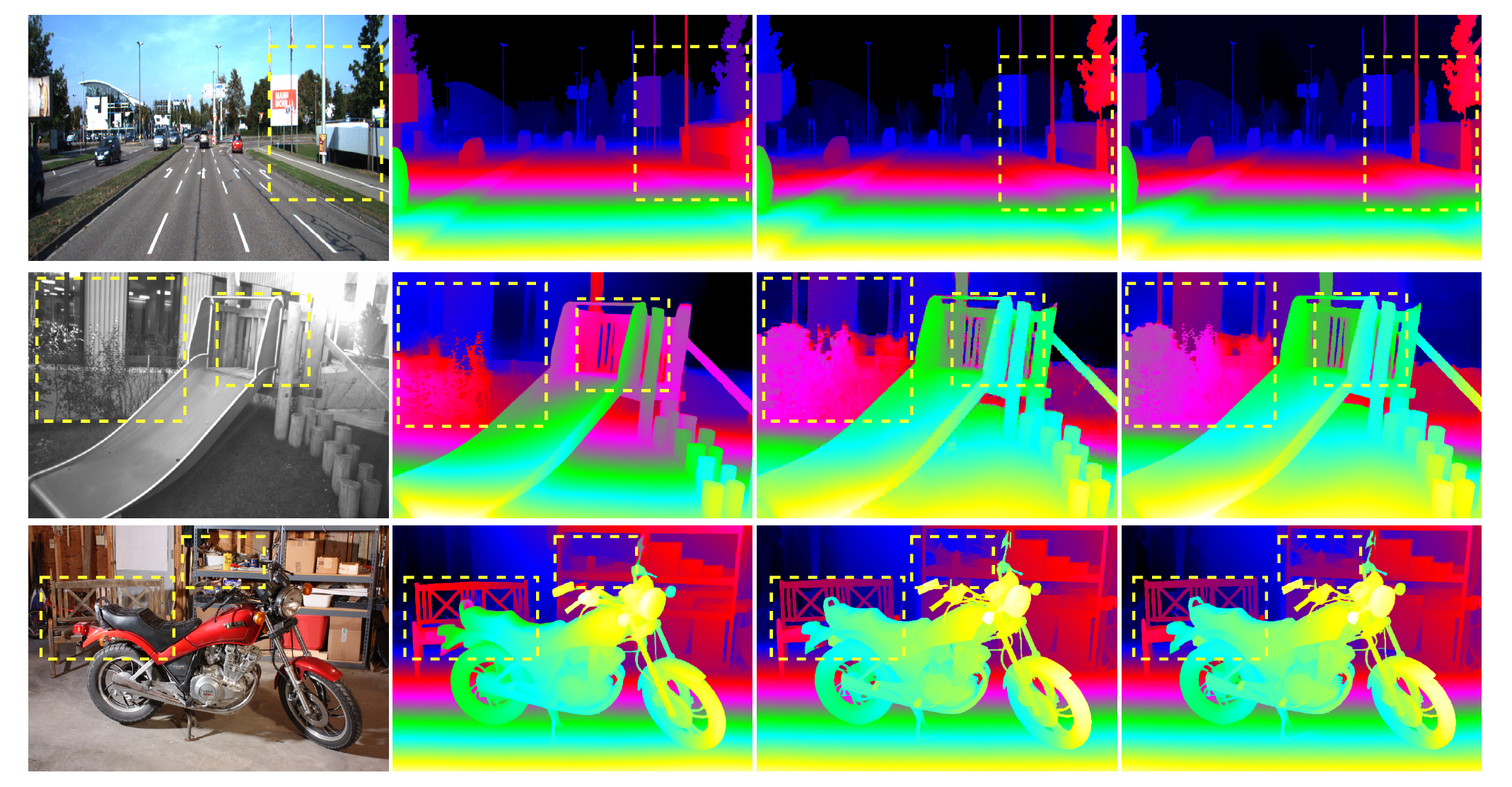}\\
  \vspace{-4pt}
  \makebox[0.245\linewidth]{\footnotesize\textsf{Input Image}}
  \makebox[0.24\linewidth]{\footnotesize\textsf{DAv2~\cite{yang2024depth}}}
  \makebox[0.24\linewidth]{\footnotesize\textsf{Ours-\emph{Mono}}}
  \makebox[0.245\linewidth]{\footnotesize\textsf{Ours-\emph{Stereo}}}\\
  \vspace{-4pt}
  \caption{\textbf{Qualitative results of zero-shot generalization} for monocular outputs, comparing with DepthAnythingV2~\cite{yang2024depth}. Our monocular depth recovers scale consistency and fine details.}
  \label{fig:mono_zero}
  \vspace{-1.5em}
\end{figure}
\subsection{Benchmark Datasets and Metric}
\noindent\textbf{Datasets.} We evaluate on five standard benchmarks spanning synthetic and real-world scenarios. 
Scene Flow~\cite{mayer2016large} provides 35454 synthetic training and 4370 testing stereo pairs ($960\times 540$ resolution) with dense, accurate disparity maps across three subsets: FlyingThings3D, Driving, and Monkaa. 
For real-world evaluation, KITTI 2012~\cite{geiger2012we} (194 training/195 testing pairs) and KITTI 2015~\cite{menze2015object} (200 training/200 testing pairs) offer driving scenes with sparse LIDAR-based disparity annotations. 
To assess generalization, we additionally include zero-shot evaluation on Middlebury 2014~\cite{scharstein2014high}, featuring high-precision structured-light disparity for indoor scenes, and ETH3D~\cite{schops2017multi}, containing mixed indoor/outdoor grayscale stereo pairs.
For qualitative analysis of challenging cases, we further test on the Booster~\cite{zamaramirez2022booster} dataset, which features indoor scenes with reflective surfaces and transparent objects to evaluate robustness under complex material properties.

\noindent\textbf{Evaluation Metrics.}
Following established protocols~\cite{menze2015object,lipson2021raft,zhang2019domaininvariant,mayer2016large}, we adopt three standard disparity evaluation metrics: (1) \textbf{EPE}: mean absolute disparity error across all valid pixels (\eg, excluding pixels with ground-truth (GT) disparity $\!>\!192$ in Scene Flow); (2) \textbf{BP-X}~\cite{wen2025foundationstereo}: proportion of pixels with absolute disparity errors exceeding $X$ pixels; and (3) \textbf{D1}: error rate where pixels fail both absolute ($>$3px) and relative ($>$5\% of GT disparity) error thresholds.

\subsection{Zero-Shot Generalization}
While monocular depth estimation (MDE) has seen notable progress in zero-shot generalization, stereo matching remains challenged by domain shifts and ambiguous regions such as reflective or transparent surfaces.

\begin{table}[!t]
\centering
\def\mywidth{0.47\textwidth} 
\resizebox{\mywidth}{!}{

\begin{tabular}{l c c c c}
\toprule
\multirow{2}{*}{Methods} & KITTI-12 & KITTI-15 & Middlebury (Q) & ETH3D \\
& (D1) & (D1) & (BP-2) & (BP-1) \\
\midrule
DSMNet~\cite{zhang2019domaininvariant} & 6.2 & 6.5 & 8.1 & 6.2\\
Mask-CFNet~\cite{rao2023masked} & 4.8 & 5.8 & - & 5.7 \\

CREStereo++~\cite{jing2023uncertainty} & 4.7 & 5.2 & - & 4.4\\
Former-RAFT-DAM~\cite{zhang2024learning} & 3.9 & 5.1 & 5.4 & 3.3 \\
RAFT-Stereo~\cite{lipson2021raft} & 4.7 & 5.5 & 9.4 & 3.3\\

IGEV-Stereo~\cite{xu2023iterative} &5.2 & 5.7 & 8.8 & 4.0\\
NMRF~\cite{guan2024neural} & 4.2 & 5.1 & 7.5 & 3.8\\
IGEV++~\cite{xu2024igev++} & 5.1 & 5.9 & 7.8 & 4.1 \\
FoundationStereo~\cite{wen2025foundationstereo} & \cellcolor{red!25}\textbf{3.2} & \cellcolor{yellow!25}4.9 & \cellcolor{yellow!25}5.5 & \cellcolor{yellow!25}1.8 \\
\midrule
BridgeDepth (Stereo) & \cellcolor{yellow!25}3.6 & \cellcolor{red!25}\textbf{4.5} & \cellcolor{red!25}\textbf{4.3} & \cellcolor{red!25}\textbf{1.3}\\
BridgeDepth (Mono$^\dag$) & 3.7 & 4.2 & 5.8 & 1.7\\
\bottomrule
\end{tabular}

}
\vspace{-0.6em}
\caption{\textbf{Zero-shot generalization evaluation.} All methods are only trained on Scene Flow data. The \colorbox{red!25}{\textbf{best}} and \colorbox{yellow!25}{second best} are marked. ($\dag$): Relative depth aligned to GT via ROE solver~\cite{wang2025moge}.
}
\label{tab:results_zero_shot}
\vspace{-1.2em}

\end{table}
\noindent\textbf{Benchmark Evaluation.}
We assess generalization capability using models exclusively trained on synthetic Scene Flow data and evaluate on the training sets of real-world datasets (KITTI, Middlebury, ETH3D) without fine-tuning.
\Cref{tab:results_zero_shot} quantitatively compares our method with current state-of-the-art approaches.
By integrating generalizable monocular cues through cross-modal fusion, our approach outperforms the baseline NMRF~\cite{guan2024neural} by 42.7\% and 65.8\% on Middlebury and ETH3D, respectively.
This dramatic improvement highlights the untapped potential of bridging geometric correspondence with vision foundation models.
It also achieves superior performance than concurrent FoundationStereo~\cite{wen2025foundationstereo} on 3/4 benchmarks.
Besides, the monocular relative depth (last row of~\cref{tab:results_zero_shot}) also exhibits good generalization performance.

\begin{figure}[!t]
  \centering
  \includegraphics[width=0.95\linewidth]{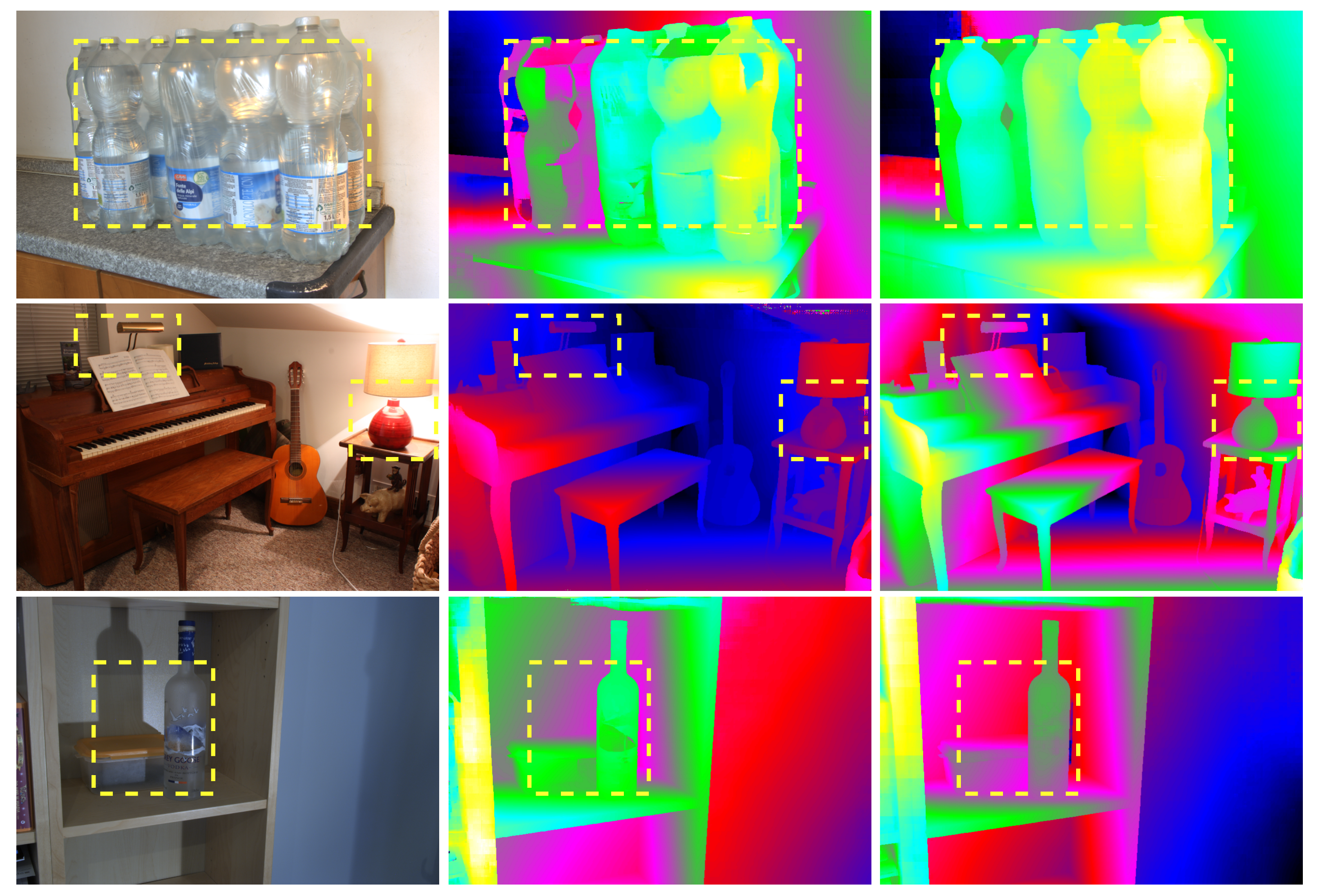}\\
  \vspace{-3pt}
  \makebox[0.33\linewidth]{\footnotesize\textsf{Input Image}}
  \makebox[0.32\linewidth]{\footnotesize \textsf{NMRF}~\cite{guan2024neural}}
  \makebox[0.32\linewidth]{\footnotesize \textsf{Our-\emph{Stereo}}}
  \vspace{-0.3em}
  \caption{\textbf{Qualitative results of zero-shot generalization} in ill-posed regions for stereo methods, comparing with NMRF~\cite{guan2024neural}.}
  \label{fig:stereo_zero}
  \vspace{-1.2em}
\end{figure}

\noindent\textbf{Generalization in Ill-posed Regions}
We assess generalization in ill-posed regions using the Booster dataset~\cite{zamaramirez2022booster}, featuring challenging indoor scenes with reflective surfaces (\eg, mirrors, glass) and transparent objects (\eg, windows, acrylic). 
Visual comparisons in \cref{fig:tsear,fig:stereo_zero} demonstrate our method overcome matching ambiguities through monocular structural priors, yielding coherent disparity estimates where matching-based methods fail. 
The NMRF baseline~\cite{guan2024neural} exhibits severe fragmentation and hallucinated boundaries, while our approach maintains structural consistency near specular highlights (\eg, left cabinet in~\cref{fig:tsear}) and transparent surfaces (\eg, the bottles in~\cref{fig:stereo_zero}). 
These results validate the benefits of combining stereo matching with monocular foundation models.
\begin{figure}[!t]
  \centering
\includegraphics[width=0.97\linewidth]{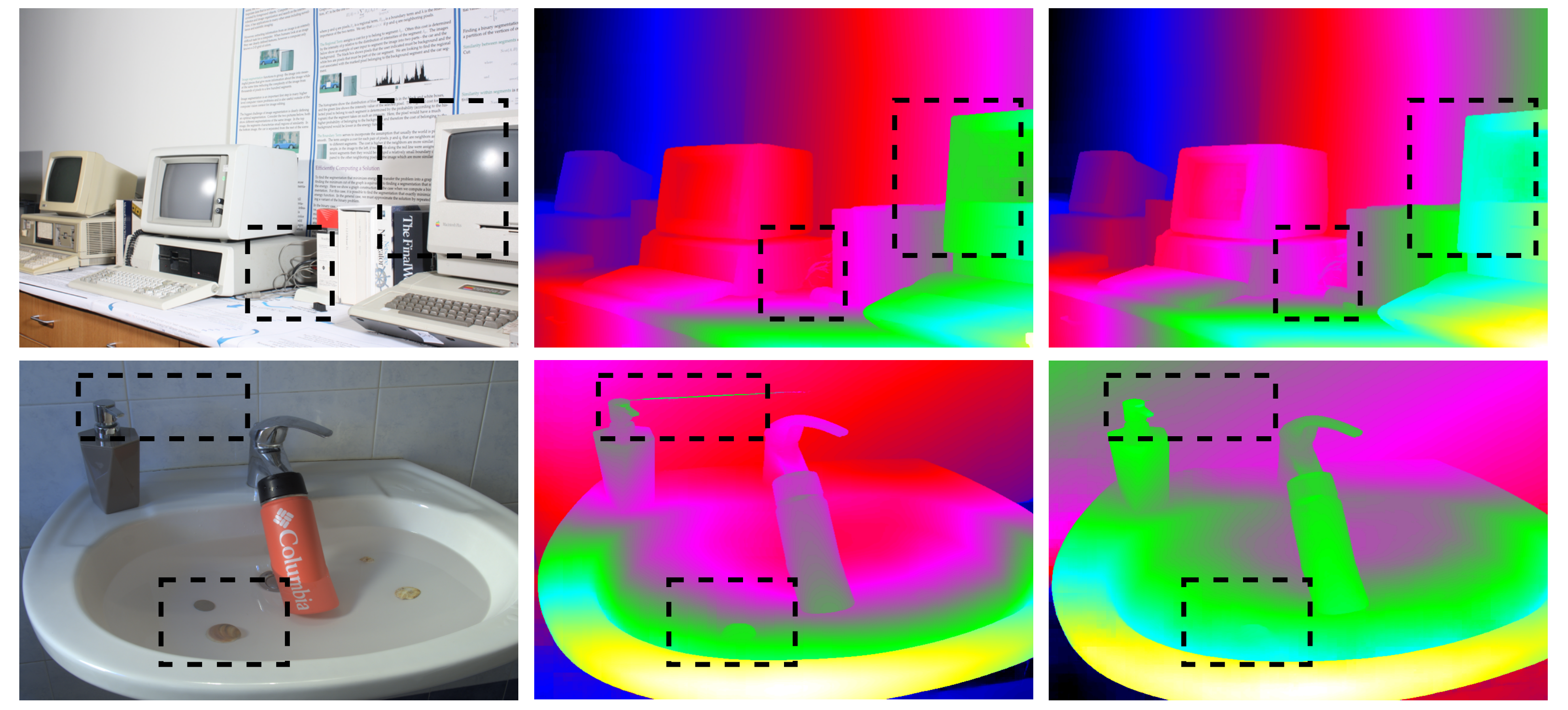}\\
\vspace{-4pt}
\makebox[0.33\linewidth]{\footnotesize\textsf{Input Image}}
\makebox[0.32\linewidth]{\footnotesize $\mathcal{L}_{\text{stereo}}$}
\makebox[0.32\linewidth]{\footnotesize$\mathcal{L}_{\text{stereo}}+\mathcal{L}_{\text{mono}}$}\\
  \vspace{-0.3em}
  \caption{\textbf{Qualitative results of dual-loss ablation study.}}
  \label{fig:abl}
  \vspace{-0.8em}
\end{figure}
\subsection{Benchmark Performance}
\textbf{Scene Flow.} As shown in~\cref{tab:results_sceneflow}, our approach achieves SOTA performance on Scene Flow test set with 3.67\% BP-1 error---18.44\% better than prior art.
Additionally, the scalability of our approach is evidenced by BridgeDepth-L, a variant with enlarged stereo feature network (details in suppl.), which achieves 3.25\% BP-1 error, demonstrating consistent gains through enhanced representation capacity.

\noindent\textbf{KITTI.} 
We adapt our Scene Flow-pretrained model to KITTI by finetuning on mixed KITTI 2012/2015 data~\cite{geiger2012we,menze2015object} for 40K iterations with $304\times 1152$ random crops, batch size 4, and reduced peak learning rate 2e-4. 
As shown in~\cref{tab:results_kitti}, our method outperforms geometry-only stereo methods (Selective-IGEV~\cite{wang2024selective}, Mocha-Stereo~\cite{chen2024motif}) and the hybrid LoS framework~\cite{li2024local}, particularly excelling in reflective regions.
~\Cref{fig:kitti_ft} confirms superior robustness for transparent windows and fine details. 
Compared to concurrent work~\cite{jiang2025defom,xu2024igev++,cheng2025monster}, our method achieves best accuracy on most metrics with significantly lower computation.

\begin{table}[!t]
\vspace{0.2em}
\small
\centering
\begin{tabular}{l c c c}
\toprule
  & ETH3D & Middlebury & Time \\
  & (BP-1) & (BP-2) & (s) \\
\midrule
baseline (NMRF-s) & 3.7 & 8.0 & 0.044\\
w/ monocular-readout & 2.3 & 5.9 & 0.123\\
w/ monocular-update & 2.1 & 5.5 & 0.128\\
w/ fine-grained & 1.6 & 4.8 & 0.147\\
w/ monocular loss & 1.3 & 4.4 & 0.147\\
\bottomrule
\end{tabular}
\vspace{-0.5em}
\caption{\textbf{Ablation study} on the effect of individual components.}
\label{tab:ablation_comp}
\vspace{-1.1em}
\end{table}
\subsection{Ablation Studies}
Our ablation studies evaluate architectural variants trained on Scene Flow, measuring zero-shot generalization (ETH3D/Middlebury accuracy) and efficiency (inference time at $960\times540$ resolution). The baseline adopts a streamlined NMRF~\cite{guan2024neural} configuration with 2 hypothesis per pixel.

As shown in~\cref{tab:ablation_comp}, the unidirectional monocular-readout yields substantial gains in stereo generalization, underscoring the critical role of monocular priors in resolving geometric ambiguities under domain shifts. 
In contrast, the monocular-update module, primarily refining the monocular branch, provides only marginal improvements to stereo accuracy.
The third row in \cref{tab:ablation_comp} illustrates that fine-grained matching cues improve not only accuracy but also stereo generalization. Our asymmetric resolution interaction provides a good tradeoff between efficiency and generalization.
The monocular loss also greatly enhances the generalization via explicit monocular-ground truth alignment supervision, which implicitly enforces the monocular-stereo consistency.
We give some qualitative examples in~\cref{fig:abl} to illustrate the efficacy of monocular loss.
We refer extended ablation studies and benchmark evaluations to the supplementary.

%% file: sec/5_conclusion.tex
\section{Conclusion}
We present a latent alignment framework that dynamically couples geometric matching with monocular contextual priors, resolving the fundamental tension between precise correspondence estimation and generalizable contextual reasoning. Validated through extensive stereo matching experiments, our bidirectional alignment mechanism substantially improves zero-shot generalization (e.g., +42.7\% on Middlebury) and robustness to ambiguous regions like reflective surfaces. 
Despite its strong performance, our implementation may compromise accuracy at high resolutions ($\ge$2K) where large disparity ranges strain the top-k hypothesis selection in the Disparity Proposal Network.
Future work could address this via a stereo feature interaction module.

\noindent\textbf{Acknowledgement.} This work is supported in part by the InnoHK initiative of the Innovation and
Technology Commission of the Hong Kong Special Administrative Region Government via the
Hong Kong Centre for Logistics Robotics, the CUHK T Stone Robotics Institute, and the HK ITF PRP under Grant PRP/038/23FX.

%% file: sec/X_suppl.tex
\def\thesection{\Alph{section}}
\setcounter{section}{0}
\setcounter{figure}{4}
\setcounter{table}{6}

\maketitlesupplementary

This supplementary material provides additional insights and evaluations to support the findings presented in our main paper.
In~\cref{supp:variant}, we elaborate on the details of the stereo feature network of BridgeDepth-L. \cref{supp:abl} presents additional ablation studies, including comparing the performance of our framework when using different monocular foundation backbones, and a comparison with a post-hoc monocular-stereo combination strategy. In~\cref{supp:res}, we report results from the official ETH3D benchmark, accompanied by a quantitative comparison between our refined monocular depth estimates and well-known monocular foundation models.

\section{BridgeDepth-L variant}
\label{supp:variant}
To assess the scalability of our approach, we introduce a variant named \textbf{OmiDepth-L}, which replaces the standard $\tt BasicEncoder$ with an enlarged stereo feature network. 
This enhanced network is built upon a pre-trained ConvNext-Tiny model~\cite{liu2022convnet}, a cutting-edge convolutional neural network recognized for its efficiency and robust feature extraction capabilities across various vision tasks.

In the BridgeDepth-L architecture, we utilize the first three ``blocks'' of the ConvNext-Tiny model.
These blocks produce hierarchical feature maps at multiple spatial resolutions: 1/4, 1/8, and 1/16 of the input image size.
This multi-scale feature extraction enables the model to capture both fine-grained local details (at higher resolutions) and broader contextual information (at lower resolutions), which are critical for high-quality correspondence estimation.
To integrate these multi-resolution features effectively, we employ a DPT fusion module.
The DPT fuses the features from different scales into a cohesive representation at 1/4 resolution with a channel dimension of 128.
Subsequently, this fused feature map is average-pooled to yield a feature representation at 1/8 resolution, which serves as the input for coarse-level inference within our bidirectional alignment pipeline.

This variant demonstrates the flexibility of our framework, as it can seamlessly scale to incorporate more powerful backbones.
By replacing $\tt BasicEncoder$ with enlarged stereo feature network, BridgeDepth-L serves as a proof-of-concept for adapting our approach to more sophisticated architectures, potentially yielding superior performance in demanding scenarios (\eg, in~\cref{tab:results_sceneflow}).

\section{Additional ablations}
\label{supp:abl}
In this section, we present additional ablation studies to further investigate the robustness and versatility of the proposed framework. These experiments focus on two key aspects: (1) evaluating the performance of our framework when integrated with different monocular foundation backbones, and (2) comparing our iterative bidirectional alignment mechanism against a post-hoc monocular-stereo combination strategy. The results of these studies validate the generalization of our method and highlight the critical role of our alignment module in achieving superior disparity estimation performance.

\subsection{Generalization with monocular backbones}
To demonstrate that our framework is not limited to a single monocular backbone (\ie, DepthAnythingV2), we test its performance using multiple state-of-the-art monocular foundation models: DepthAnythingV2-L (DAv2-L)~\cite{yang2024depth}, DepthAnythingV2-B (DAv2-B)~\cite{yang2024depth}, UniDepthV2 (UDv2)~\cite{piccinelli2025unidepthv2}, and MoGe~\cite{wang2025moge}.
Each backbone is integrated into our framework, and their performance is evaluated using in-domain and zero-shot experiments.
All models are trained exclusively using Scene Flow data.
The results, summarized in , indicate that the monocular backbone indeed affects generalization to some extent, our method exhibits significant improvements over baseline under all backbones. Thus, generality is validated.

\begin{table}[t]
\footnotesize
\caption{\textbf{In-domain accuracy on Scene Flow test set and zero-shot generalization on well-known real-world datasets.}}
\label{supp:tab_abl}
\vspace{-0.5em}
\renewcommand{\thetable}{{\alph{table}}}
  \centering
  \scalebox{0.75}{
  \begin{tabular}{c|c|cccc}
    \toprule
    \multirow{2}{*}{Backbone} & SceneFlow & KITTI-12 & KITTI-15 & ETH3D & Middlebury (Q)\\
    & EPE $\downarrow$ & D1 $\downarrow$ & D1 $\downarrow$ & BP-1 $\downarrow$ & BP-2 $\downarrow$\\
    \midrule
    baseline & 0.47 & 4.5 & 5.2 & 3.7 & 8.0\\
    \rowcolor{lightgray}
    DAv2-B & 0.37 & 3.8 & 4.7 & 1.4 & 4.7\\
    DAv2-L & 0.37 & 3.7 & 4.5 & 1.3 & 4.4\\
    UDv2 & 0.36 & 3.7 & 4.8 & 1.8 & 5.1\\
    MoGe & 0.36 & 3.3 & 4.5 & 1.8 & 5.1\\
    \midrule
    \rowcolor{lightgray}
    DAv2-B (post-hoc)&0.44&4.1&4.8&3.2&7.4\\
    \bottomrule
    \end{tabular}
  }
  \vspace{-1.4em}
\end{table}

\subsection{Post-hoc fusion}
We also compare our iterative bidirectional alignment mechanism with a post-hoc fusion strategy.
In our proposed framework, monocular contextual features are iteratively aligned with stereo hypothesis volumes during the stereo aggregation process. 
In contrast, the post-hoc baseline directly fuses the monocular features with the aggregated stereo hypothesis volume using a lightweight ResNet, bypassing the iterative alignment step. 
For this experiment, we employ the ViT-B version of DAv2 as the monocular backbone.
The performance of the post-hoc fusion strategy is reported in the last row of Table~\ref{supp:tab_abl}. While this baseline yields some improvement over purely monocular or stereo methods, it underperforms compared to our iterative alignment approach. This gap in performance suggests that the primary gains in our framework arise from the dynamic, bidirectional alignment of monocular and stereo cues, rather than a static fusion of features. By enabling mutual refinement between modalities, our method produces more accurate and reliable correspondence estimates.

\section{Additional results}
\label{supp:res}

In this section, we report the performance of our method on the official ETH3D benchmark, a widely adopted dataset designed for evaluating stereo matching methods. The benchmark is renowned for its challenging scenes, encompassing both indoor and outdoor environments with intricate geometries, textureless regions, and reflective surfaces. Our approach achieves state-of-the-art (SOTA) performance on this benchmark, underscoring its ability to deliver robust and accurate depth estimates across diverse conditions.

\begin{table}[t]
\footnotesize
\caption{\textbf{Results on ETH3D leaderboard (test set)}}
\label{supp:tab_eth3d}
\vspace{-1.0em}
\renewcommand{\thetable}{{\alph{table}}}
  \centering
  \begin{tabular}{c|cccc}
    \toprule
    Method & BP-1 & BP-2 & AvgErr & RMS \\
    \midrule
    LoS~\cite{li2024local} & 1.03 & 0.32 & 0.15 & 0.34\\
    IGEV++~\cite{xu2024igev++} & 1.58 & 0.76 & 0.19 & 0.74\\
    Selective-IGEV~\cite{wang2024selective} & 1.56 & 0.51 & 0.15 & 0.57\\
    DEFOM-Stereo~\cite{jiang2025defom} & 0.78 & \underline{0.19} & \textbf{0.11} & \textbf{0.26}\\
    FoundationStereo~\cite{wen2025foundationstereo} & \textbf{0.48} & 0.26 & 0.13 & 0.61\\
    BridgeDepth (Ours) & \underline{0.50} & \textbf{0.16} & \textbf{0.11} & \textbf{0.26}\\
    \bottomrule
    \end{tabular}
  \vspace{-1.0em}
\end{table}

We also present a quantitative comparison between our refined monocular depth estimates and those generated by well-known monocular foundation models. This comparison, which will be detailed in~\cref{supp:tab_depth}, highlights the superior accuracy of our method. The key to this improvement lies in the integration of geometric precision from stereo data, which introduces additional constraints absent in purely monocular approaches. By leveraging stereo information, our refined monocular depth estimates gain the geometric consistency and precision inherent in stereo vision, resulting in significantly enhanced depth maps.

\begin{table}[t]
\footnotesize
\caption{\textbf{Monocular evaluation.}}
\label{supp:tab_depth}
\vspace{-1.2em}
\renewcommand{\thetable}{{\alph{table}}}
  \centering
  \scalebox{0.55}{
    \begin{tabular}{c|cc|cc|cc|cc}
    \toprule
    \multirow{2}{*}{Model}& \multicolumn{2}{c|}{KITTI-12} & \multicolumn{2}{c|}{KITTI-15} & \multicolumn{2}{c|}{ETH3D} & \multicolumn{2}{c}{Middlebury}\\
    & Abs Rel $\downarrow$ & $\delta<1.25 \uparrow$ & Abs Rel $\downarrow$ & $\delta<1.25 \uparrow$ & Abs Rel $\downarrow$ & $\delta<1.25 \uparrow$ & Abs Rel $\downarrow$ & $\delta<1.25 \uparrow$ \\
    \midrule
    DAv2-L & 0.093 & 0.925 & 0.084 & 0.933 & 0.055 & 0.965 & 0.081 & 0.943\\
    UDv2 & 0.054 & 0.967 & 0.067 & 0.949 & 0.052 & 0.972 & 0.056 & 0.982\\
    Ours & \textbf{0.033} & \textbf{0.982} & \textbf{0.048} & \textbf{0.975} & \textbf{0.020} & \textbf{1.000} & \textbf{0.019} & \textbf{0.990}\\
    \bottomrule
    \end{tabular}
  }
  \vspace{-1.2em}
\end{table}